\documentclass{article}
\usepackage{ijcai18}

\usepackage{times}
\usepackage{soul}
\usepackage{url}
\usepackage[hidelinks]{hyperref}
\usepackage[utf8]{inputenc}
\usepackage[small]{caption}
\usepackage{graphicx}
\usepackage[linesnumbered,ruled,vlined]{algorithm2e}
\usepackage{epstopdf}
\usepackage{amssymb}
\usepackage{paralist}

\newcommand{\bigO}{\mathcal{O}}
\newcommand{\Dataset}{\mathcal{D}}

\newcommand{\OPEN}{\textsc{Open}}
\newcommand{\CLOSED}{\textsc{Closed}}	

\title{Safe learning-based optimal motion planning for automated driving}

\author{
Zlatan Ajanovi{\'{c}}$^1$,
Bakir Lacevi{\'{c}}$^2$,
Georg Stettinger$^1$,
Daniel Watzenig$^{1,3}$,
Martin Horn$^3$
\\ 
$^1$ Virtual Vehicle Research Center, Graz, Austria\\
$^2$ University of Sarajevo, Sarajevo, Bosnia and Herzegovina\\
$^3$ Graz University of Technology, Graz, Austria\\
\{zlatan.ajanovic, georg.stettinger, daniel.watzenig\}@v2c2.at,
bakir.lacevic@etf.unsa.ba,
martin.horn@tugraz.at   
}

\begin{document}

\maketitle

\begin{abstract}
  This paper presents preliminary work on learning the search heuristic for the optimal motion planning for automated driving in urban traffic. 
  Previous work considered search-based optimal motion planning framework (SBOMP) that utilized numerical or model-based heuristics that did not consider dynamic obstacles. Optimal solution was still guaranteed since dynamic obstacles can only increase the cost. However, significant variations in the search efficiency are observed depending whether dynamic obstacles are present or not. 
  This paper introduces machine learning (ML) based heuristic that takes into account dynamic obstacles, thus adding to the performance consistency for achieving real-time implementation.
  
\end{abstract}

\section{Introduction}

Vehicle automation is based on classic robotics Sensing-Planning-Acting cycle, where Motion Planning (MP) is the crucial step. Many different approaches for MP are available \cite{schwarting2018review} but still, finding a collision-free motion plan, while taking into account system dynamics, dynamic obstacles and possibly desired criteria (cost function) in a real time is an unsolved challenge.

Beside environment perception, where Deep Learning (DL) proved to be very useful, it was not extensively used for other vehicle automation tasks such as vehicle motion planning and control. So far, ML was used to learn longitudinal driving (e.g. energy efficient \cite{gaier2014evolving}) and achieved comparable energy-efficiency as search based solutions, with improved  computational performance. For lateral driving, DL was used to imitate the human driver with the goal to only keep the vehicle on the road \cite{bojarski2016end}. These situations are rather simple compared to driving in a complex, dynamic urban environment. Several works used reinforcement learning to learn driving in a dynamic environment \cite{shalev2016safe,fridman2018deeptraffic}, but the approaches consider simple models and the results are not close to optimal.

Successes of DL in robotic manipulation \cite{levine2016end} and playing the game of Go \cite{silver2017mastering} as well as approximation of planning modules by network \cite{tamar2016value} provide arguments for considering DL for vehicle motion planning too.

\begin{figure}
  \includegraphics[trim= 32 10 -12 75,clip, width=\linewidth]{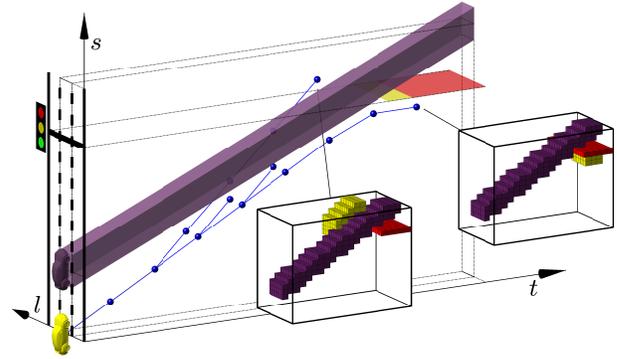}
  \setlength{\abovecaptionskip}{-7pt}
  \setlength{\belowcaptionskip}{-10pt}
  \caption{\small MP for ego vehicle (yellow) in scenario with receding vehicle (purple) and traffic light. Situation representation for 2 nodes.}
  \label{fig:sit_rep}
\end{figure}

The proposed approach utilizes ML to enhance computational performance of deterministic MP while keeping guarantees on optimality and transparency. The main contribution of this work can be summarized as: 

\begin{inparaenum}[i)]
	\item systematic dataset generation from exact optimal solutions for supervised learning of optimal behavior, 
	\item convenient representation of driving situation as input for machine learning algorithm,
	\item use of machine learning for a heuristic in a deterministic planning framework with guaranteed maximum deviation from optimal solution,
	\item use of receding horizon approach instead of greedy policy search.
\end{inparaenum}

\section{Problem statement}
The main problem considered in this work is inconsistent and potentially unsatisfactory performance of deterministic motion planning approaches, in particular Search Based Optimal Motion planning framework (SBOMP) \cite{ajanovic2018search}. As heuristics used in SBOMP (numerical $h_{DP}$ \cite{myBookCh2017,myAMAA2017} and model based $h_{MB}$ \cite{myCTS2017}) did not consider dynamic obstacles, significant computational performance variations have been observed depending whether dynamic obstacles are present or not. 
Performance is measured by the number of nodes explored. 

To integrate dynamic obstacles in heuristic function ML techniques will be used in this work.

\section{Learning-based optimal motion planning}
The original SBOMP for optimal motion planning is enhanced by ML heuristic  $h_{ML}$, bounded by  $h_{DP}$ for providing guarantees on sub-optimality. Proposed $h_{ML}$ takes as input a 3D structure representing node $n$ and a driving situation (obstacles $\bigO$) and returns as a result a scalar value representing an estimated cost to reach horizon limits from that node. Using modified SBOMP and  DP heuristic ($h_{DP}$), dataset ($\Dataset$) of exact  input-output data points is generated. Dataset is then used for training of the ML heuristic. In principle, this approach differs from reinforcement learning since it is supervised and from imitation learning since the exact optimal solution is used instead of expert demonstrations.

\subsection{Situation representation}

Node $n$ and driving situation are represented by a discretized 3D structure with lane $l$, longitudinal distance $s$ and time $t$ dimensions (Fig. \ref{fig:sit_rep}). 3D structure has $N_{ks\mathrm{hor}} \times N_{kt\mathrm{hor}} \times N_{kdl}$ cells. Each cell represents a part of the search space and can be free or occupied by some obstacle or ego vehicle. Obstacles are uniquely marked for several types: other vehicles, traffic lights, forbidden lane change, etc. as shown in \cite{ajanovic2018search}. Parent node $n$ is represented by virtual obstacle as if the vehicle is continuing to move with velocity $n.v$ from it's initial position. The advantage of this formulation is that the representation keeps consistent 3D size for every driving situation regardless of the number of obstacles and it is computed only once per replanning instance.

\subsection{Dataset generation}
For generation of dataset $\Dataset$, the SBOMP framework was modified to enable theoretically inexhaustible generation from different scenarios and initial conditions as it is shown in Algorithm \ref{alg3}. Contrary to the original SBOMP, where the goal was to find only one collision-free trajectory, for dataset generation the goal is to generate as many different data points as possible. Therefore, the search is executed not only until horizon is reached by some trajectory, but until all nodes in $\OPEN$ list are explored. Trajectory branches are constructed backward from each node $n_h$ at the end of horizons, starting from nodes which reached horizon first. For each node $n$ on the branch, corresponding 3D structure and cost are stored in dataset $\Dataset$. Cost is computed by subtracting $g(n)$ value from the cost of the final node $g(n_h)$. It can be noted that calculated cost represents \textit{cost-to-horizon} and not \textit{cost-to-go} anymore, but this should not affect the results of the search. The remaining nodes, not belonging to the branches that reach horizons, lead to the collision, and thus are assigned with infinite cost. These nodes partially resemble inevitable collision state \cite{fraichard2004inevitable}. 
\setlength{\textfloatsep}{0pt}
\begin{algorithm}[t]
\DontPrintSemicolon
\fontsize{8pt}{9pt}\selectfont
	\SetKwData{n}{$n$}
	\SetKwFunction{Expand}{Expand}
	\SetKwInOut{Input}{input}\SetKwInOut{Output}{output}
	\Input{$k_{max}$\tcp*[r]{Number of initial poses}	} 
	\Output{$\Dataset$\tcp*[r]{Dataset}	} 
	\BlankLine
	\Begin{
		\ForEach{$k \in [1, k_{max}]$}{
 			$n  \gets rand()$\tcp*[r]{random initial pose}
 			$\bigO \gets rand()$\tcp*[r]{random driving scenario}
			$\CLOSED \gets \varnothing$\tcp*[r]{list of closed nodes}
			$\OPEN \gets n $\tcp*[r]{list of opened nodes}
			\While{$\OPEN \neq \varnothing $}{
			 			
				$\CLOSED\gets \CLOSED \cup n$ \tcp*[r]{Exploring}
				$\OPEN \gets \OPEN \setminus n$\;		
				\ForEach{ $n' \in$  \Expand{$n,\bigO$, $h(s, v)$}} {
					\uIf{ $n' \in \CLOSED$}{
						$\textit{continue} $\;
					}
					\uElseIf{$n' \in\OPEN$}{
						\uIf{ $\textit{new} $ $n'$  $\textit{is better}$
						}{
							$n'.parent \gets n$ \tcp*[r]{update parent}
						}
						\Else{
							$\textit{continue} $\;
						}}
	 				\Else { 
						$\OPEN \gets \OPEN \cup n' $ \tcp*[r]{add to list}					
					}
				}
				$n \gets \textit{argmin } n.f \in \OPEN$ \;
			}		
			$\Dataset \gets \varnothing$ \tcp*[r]{Dataset}
			\ForEach{ $n_h \in \CLOSED  \mid n_h.s = S_{hor} \lor n_h.t = T_{hor}$} {			
				$n \gets n_h$ \tcp*[r]{end of one traj. branch}
				\While{ $n \in \CLOSED$} {
					$\Dataset \gets \Dataset \cup \{n, \bigO(n),  n_h.g - n.g\}$\;
					$\CLOSED \gets \CLOSED \setminus n$\;
					$n \gets n.parent$\;
				}
			}
			\ForEach{ $n \in \CLOSED$} {
			
			$\Dataset \gets \Dataset \cup \{n, \bigO(n), inf\}$ \tcp*[r]{dead-end trajs}
			}
			}
		\Return{$\Dataset$}\;
	}
\caption{modified SBOMP for dataset generation\label{alg3}}
\end{algorithm}

\subsection{ML Heuristic}

Proposed ML heuristic is bounded by DP heuristic (\ref{eq1}) so that guarantees on sub-optimality can be provided. In this way, heuristic is $\varepsilon$-admissible so the solution is always maximum $\varepsilon$ times greater than the optimal solution \cite{likhachev2004ara}. Values of $\varepsilon$ closer to $1$ guarantee smaller deviation from optimal solution but reduce computational performance.

\begin{equation} \label{eq1}
h_{DP} \leq h_{ML} \leq \varepsilon \cdot h_{DP}, \quad \varepsilon \geq 1. 
\end{equation}

\section{Conclusion and outlook}
The presented approach offers the possibility to include Machine Learning into deterministic motion planning framework, promising significant performance improvements manifested in reduced number of nodes explored compared to those obtained using numerical ($h_{DP}$) and model based ($h_{MB}$) heuristics while keeping guarantees on sub-optimality of the solution. Appropriate ML architecture for this problem have yet to be developed and learnability validated.

\section*{Acknowledgments}

The project leading to this study has received funding from the European Union’s Horizon 2020 research and innovation programme under the Marie Skłodowska-Curie grant agreement No 675999, ITEAM project.\par
VIRTUAL VEHICLE Research Center is funded within the COMET – Competence Centers for Excellent Technologies – programme by the Austrian Federal Ministry for Transport, Innovation and Technology (BMVIT), the Federal Ministry of Science, Research and Economy (BMWFW), the Austrian Research Promotion Agency (FFG), the province of Styria and the Styrian Business Promotion Agency (SFG). The COMET programme is administrated by FFG.\par

\bibliographystyle{named}
\bibliography{./literature/library}

\end{document}